\documentclass{article}

\usepackage[utf8]{inputenc} 
\usepackage[T1]{fontenc}    
\usepackage{hyperref}       
\usepackage[a4paper]{geometry}
\usepackage{url}            
\usepackage{booktabs}       
\usepackage{amsfonts}       
\usepackage{nicefrac}       
\usepackage{microtype}      
\usepackage{lipsum}
\usepackage{amsmath}
\usepackage{upgreek}

\title{Characterizing the invariances of learning algorithms using category theory}

\author{
  Kenneth D. Harris \\
  University College London\\
  London WC1E 6BT, UK \\
  \texttt{kenneth.harris@ucl.ac.uk} \\
}
\usepackage{bm}
\usepackage{tikz-cd}
\usepackage{amscd}
\begin{document}
\maketitle


\newcommand{\cat}[1]{{\bm{\mathsf{#1}}}}
\newcommand{\cX}{\cat{X}}
\newcommand{\cY}{\cat{Y}}
\newcommand{\cI}{\cat{I}}
\newcommand{\cSet}{\cat{Set}}
\newcommand{\cFin}{\cat{FinVec}}
\newcommand{\cEuc}{\cat{Euc}}
\newcommand{\bR}{\mathbb{R}}
\newcommand{\vx}{\mathbf{x}}
\newcommand{\vy}{\mathbf{y}}
\newcommand{\vf}{\mathbf{f}}
\newcommand{\va}{\mathbf{a}}
\newcommand{\veta}{\boldsymbol{\upeta}}
\newcommand{\vxi}{\boldsymbol{\upxi}}

\begin{abstract}
Many learning algorithms have invariances: when their training data is transformed in certain ways, the function they learn transforms in a predictable manner. Here we formalize this notion using concepts from the mathematical field of category theory. The invariances that a supervised learning algorithm possesses are formalized by categories of predictor and target spaces, whose morphisms represent the algorithm's invariances, and an index category whose morphisms represent permutations of the training examples. An invariant learning algorithm is a natural transformation between two functors from the product of these categories to the category of sets, representing training datasets and learned functions respectively. We illustrate the framework by characterizing and contrasting the invariances of linear regression and ridge regression.
\end{abstract}

\section{Introduction}

To train a supervised learning algorithm, one supplies a set of training examples: a set of predictor variables $x_i$ lying in some space $X$, and a matched set of target variables $y_i$ in some space $Y$. The algorithm outputs a function $f: X \to Y$, such that for all $i$, $y_i$ is well approximated by $f(x_i)$.

Many learning algorithms show invariances. By this we mean that there are certain ways one can transform the training data that cause the learned function to transform in a predictable manner. For example, rotating the target variables in the training set, could cause the learned function to rotate accordingly. If we want to know what function an invariant algorithm would have learned if it had been trained on transformed data, we don't need to run it again: we can just transform the learned function appropriately. The types of input transformations for which this is possible define the invariances of the algorithm. 

This note formalizes how to describe the invariances of learning algorithms using concepts from category theory. We illustrate the formalism by characterizing the invariances of two simple and widely-used learning algorithms: multivariate linear regression and ridge regression. Our conclusions formalize and generalize an intuition that users of these algorithms already know: when applying linear regression, different predictor variables can have arbitrary scales; but when using ridge regression the scales of all predictors should be comparable.

Although category theory is often seen as a difficult area of pure mathematics, the ideas formalized here are at heart simple, and will be already intuitively understood by most machine learning practitioners. We aim to formalize these ideas in a way that will be understandable by statisticians and machine learning practitioners who have no prior knowledge of category theory. Category theory has been applied to statistics and machine learning before \cite{mccullagh2002statistical, fong2017backprop, culbertson2013bayesian}, and invariance of data representations has been discussed in \cite{anselmi2016invariance}, however to our knowledge the current framework has not yet been described. Introductions to category theory for applied scientists can be found in \cite{spivak2014category,fong2019invitation}, and for mathematicians in \cite{riehl2017category, mac2013categories}.

\section{Learning algorithms are natural transformations}

To formalize the types of invariances a learning algorithm must respect, we will consider the spaces $X$ and $Y$ of predictor and target variables, as objects in \emph{categories} $\cX$ and $\cY$. A category specifies two things: the types of spaces the algorithm can accept as input (the \emph{objects}), and the type of invariances it must respect (the \emph{morphisms}). For our example of linear regression, the predictors and targets are finite-dimensional vectors, so the objects in the category are vector spaces of different dimensions. The category also specifies the types of invariance the algorithm must respect, by defining a set of transformations between objects: these are the morphisms. For example, if the algorithm must respect any linear transformation of its inputs, the morphisms consist of all linear transformations; this defines the full category of finite-dimensional vector spaces, termed $\cat{FinVec}$. If instead the algorithm need only respect rotations (i.e. orthogonal transformations), rather than any linear maps, this defines a category of Euclidean spaces $\cEuc$. If the algorithm must respect a any invertible linear map, even if not orthogonal, this defines an intermediate category termed $\cat{FinVec}_{iso}$. By specifying the categories $\cX$ and $\cY$, we are defining the types of data an algorithm takes for as predictors and targets (defined by the objects), and the types of invariances we require the algorithm to posses (defined by the morphisms). The more morphisms that the category contains, the more invariances will be required.

Many learning algorithms are invariant to permuting the order of the training examples. For example this is true of algorithms trained in batch mode, although not for algorithms trained in online mode. To formalize this idea, we define another category $\cI$ that indexes the training examples. Most often, the objects of this category will be finite sets, and the morphisms will be permutations of these sets (i.e. bijective functions). This category is termed $\cSet_{iso}$. If we do not require the algorithm to be invariant under permutations of the training set, we instead define $\cI$ to be a \emph{discrete category}, that has no morphisms other than the identity.

Next, we must formalize how our morphisms affect the full training data set, and how we want them to affect the algorithm's output. Do do so we require a construction known as a \emph{functor}. We will define two functors $D$ and $P$, that describe the way that training datasets, and output functions, should change under morphisms of the training data. These functors do not define a particular training dataset, or a particular output function, or a particular learning algorithm. Instead, they lay out the "rules of the game": they define the possible training datasets and possible output functions, and how we would like them to transform under morphisms of our input spaces. The functors therefore summarize what it means for a learning algorithm to be invariant. We can always define the functors, but that does not guarantee there are any non-trivial learning algorithms that respect their rules; that is something we have to show on a case-by-case basis.

Our first functor $D$ describes the set of possible training datasets. In technical language, $D$ is a functor from the category $\cX \times \cY \times \cI^{op}$ to the category of sets. What this means is that given objects $X\in \cX$, $Y\in \cY$ and $I \in \cI$, the functor defines a set $D(X,Y,I)$ containing all possible training datasets the algorithm could be given. Every dataset in $D(X,Y,I)$ is therefore a set of the form $\{(x_i, y_i): i \in I\}$. The functor $D$ also specifies how this dataset should transform under morphisms of $X$, $Y$, or $I$. They transform in the obvious way. Given a morphism $\xi:X\to X'$ (e.g. a rotation of the predictor variables), the training dataset transforms to  $\{(\xi(x_i), y_i): i \in I\}$. Given a morphism $\eta:Y\to Y'$ of the training targets, it transforms to  $\{(x_i, \eta(y_i)): i \in I\}$; and given a permutation of the input indices $\sigma:I'\to I$, it transforms to  $\{(x_{\sigma(i)}, y_{\sigma^(i)}): i \in I'\}$. Note we have allowed the possibility that $\xi$ and $\eta$ send the training data into different spaces $X'$ and $Y'$, for example a vector space of different dimension; the language of category theory allows us to define invariances under these kinds of morphisms also. 

Our second functor $P$ describes the set of possible output functions the algorithm could produce, and the way we would like these outputs to transform under morphisms. $P$ is a functor from $\cX^{op} \times \cY \times \cI^{op}$ to the category of sets (the symbol $^{op}$ represents contravariance, as explained below). Given objects $X\in \cX$ and $Y \in \cY$, $P(X,Y)$ represents the set of functions from $X$ to $Y$ that the algorithm could output. This set does not depend on the index object $I$: it is independent of the number of training examples. For example, if $X=\bR^p$ and $Y=\bR^q$, then $P(X,Y)$ will be the set of functions from $\bR^p$ to $\bR^q$. For the linear regression examples below, these will be linear functions, but this need not be the case in general; for example support vector machine or neural network classifiers can learn nonlinear functions between vector spaces. 

The functor $P$ also specifies how we would like a function $f$ to change following a morphism of the training set. Given a morphism of the target space $\eta:Y\to Y'$, we would like the function $f$ to transform to the chained function $f' = \eta \circ f$. In other words, if $f(x)=y$, and $\eta(y)=y'$, then $f'(x)=y'$; the function matches the transformation of its training targets. To describe how $f$ needs to change under a morphism of the predictor space $\xi:X'\to X$ involves a subtlety known as \emph{contravariance}. If $\xi(x') = x$, we now want $f'(x') = f(x)$; in other words $f' = f \circ \xi$.  Finally, it is easy to describe how $f$ should change under allowed morphisms of the index set $I$: it shouldn't change at all. 


We are now ready to define what we mean by an invariant learning algorithm: a \emph{natural transformation} from $D$ to $P$. A natural transformation is a family of mappings, one for each object in the source category, that respect the category's morphisms. Learning algorithms are indeed families: linear regression, for example, defines an algorithm for every possible dimension of predictor and target space, and for every possible number of input examples. Formally, for any predictor space $X \in \cX$, any target space $Y \in \cY$, and any index set $I\in \cI$, a learning algorithm $\alpha$ provides a mapping $\alpha_{X,Y,I}$ from the set of possible datasets $D(X,Y,I)$ to the set of possible learned functions $P(X,Y)$. To count as an invariant algorithm, it needs to satisfy a condition of naturality on our three categories $\cX$, $\cY$, and $\cI$.

The invariance of the algorithm under morphisms $\eta : Y\to Y'$ is summarized by  the condition that if $f=\alpha_{X,Y,I}(\{(x_i, y_i): i \in I\})$ and $f'=\alpha_{X,Y',I}(\{(x_i, \eta(y_i)): i \in I\})$, then $f'(x) = \eta(f(x))$. This can be summarized graphically by saying that the following diagram "commutes", i.e. that whether we follow the top and right arrows or the bottom and left arrows, we get the same result: 

$$
\begin{CD}
D(X,Y,I) @>{D(1_X, \eta, 1_I)}>> D(X,Y',I) \\
@V\alpha_{X,Y,I} VV @VV\alpha_{X,Y',I} V \\
P(X,Y) @>P(1_X,\eta)>> P(X,Y')
\end{CD}
$$

In the above diagram, the notation $D(1_X, \eta, 1_I)$ means the rule defined by the functor $D$ for how datasets change when the training targets transform as $y_i \mapsto \eta(y_i)$.
Similarly, the notation $P(1_X, \eta)$ defines how the functor $P$ requires the learned function $f$ to change under a morphism of $Y$: $f\mapsto \eta \circ f$. 

An invariant learning algorithm should also respect morphisms of the predictor space $X$. This time, because of the contravariant dependence on $X$, invariance requires a slightly different condition known as a \emph{dinatural transformation}. Given a morphism $\xi : X'\to X$, this is the condition that if $f'=\alpha_{X',Y,I}(\{(x_i', y_i): i \in I\})$ and $f=\alpha_{X,Y,I}(\{(\xi(x_i'), y_i): i \in I\})$, then $f'(\xi(x)) = f(x)$. Again, this can be summarized by a commutative diagram:

$$
\begin{CD}
D(X',Y,I) @>{D(\xi, 1_Y, 1_I)}>> D(X,Y,I) \\
@V\alpha_{X',Y,I} VV @VV\alpha_{X,Y,I} V \\
P(X',Y) @<P(\xi, 1_Y, 1_I)<< P(X,Y)
\end{CD}
$$

Finally, we require that the algorithm transform appropriately under morphisms of the index object $I$. Given a morphism $\sigma: I'\to I$, invariance requires that the function $f=\alpha_{X,Y,I}(\{(x_{\sigma(i)}, y_{\sigma(i)}): i  \in I' \})$ is the same the original $f'=\alpha_{X,Y,I'}(\{(x_i, y_i): i \in I'\})$. This notion of invariance is stronger than invariance in $X$ and $Y$: it requires not just a predictable transformation, but equality $f=f'$. Generally, we will consider $\sigma$ to be a permutation $I\to I$, and invariance under this means that it makes no difference what order samples are presented in. However, we have again allowed a mapping from a different index object $I'$ (note that the morphism again goes from $I'\to I$, indicating contravariant dependence on $I$); this allows extensions such as taking $\sigma$ to be a 2-to-1 mapping, for which invariance would mean that training on double the dataset makes no difference. We can summarize invariance in $I$ in another commutative diagram:

$$
\begin{CD}
D(X,Y,I') @>{D(1_X, 1_Y, \sigma)}>> D(X,Y,I) \\
@V\alpha_{X,Y,I'} VV @VV\alpha_{X,Y,I} V \\
P(X,Y) @= P(X,Y)
\end{CD}
$$




\section{Linear regression}

We now use this framework to characterize the invariances multivariate linear regression. Our predictor and target spaces $X$ and $Y$ are both finite-dimensional real vector spaces. We will characterize the invariances of linear regression by saying which precise categories of vector spaces they can come from, if the learning algorithm is to be a natural transformation.

Linear regression finds the linear map $f:X\to Y$ that minimizes the sum-squared error function:
\begin{equation}
    \label{eq:linregerr}
    E = \sum_{i \in I} ||{y_i - f(x_i)}||^2
\end{equation}

There is an exact solution to this problem \cite{friedman2001elements, anderson1984introduction}. Consider the problem of predicting $q$-dimensional target vectors from $p$-dimensional predictors, with $N$ training examples; so $X=\bR^p$, $Y=\bR^q$, and $I=\{1\ldots N\}$. Concatenate the predictor examples $\{x_i\}$ in a $N \times p$ matrix $\vx$, and the target examples $\{y_i\}$ in a $N \times q$ matrix $\vy$. Then, provided $\vx$ has rank $p$ (i.e. provided the vectors $x_i$ span $X$), the optimal predictor is the linear map represented by the matrix 
\begin{equation}
\label{eq:linregsol}
\vf= (\vx^T \vx)^{-1} \vx^T \vy. 
\end{equation}

If $\vx$ has rank $<p$ then $\vx^T \vx$ is not invertible, and the problem is underconstrained: there are infinitely many solutions $\vf$ that all have the same minimum error. We will return to this possibility later, but for now assume $\vx$ has rank $p$ so there is a unique optimal solution $\vf$.

Consider how the linear regression output $f$ transforms under an arbitrary morphism $\eta:Y\to Y'$. If $Y'$ has dimension $r$, we can represent this map by a $q \times r$ matrix $\veta$, that sends the training targets $\vy$ to $\vy \veta$. From equation (\ref{eq:linregsol}), we see the output will then transform as $\vf \mapsto \vf \veta$; in other words, the naturality condition is satisfied for any linear map $\eta$, so linear regression is invariant with $\cY = \cat{FinVec}$.

Next consider how $f$ transforms under an invertible map $\xi:X\to X$. Clearly, if we change $x_i\mapsto \xi(x_i)$ and $f\mapsto f \circ \xi^{-1} $, then the error function (\ref{eq:linregerr}) is unchanged. Because there is only one solution with this error when $rank(\vx)=p$, and because invertible maps preserve ranks, we see that $f \circ \xi^{-1} $ is the optimal solution following this transform. Alternatively, we could see this by replacing $\vx$ by $\vx \vxi$ in equation (\ref{eq:linregsol}). We therefore conclude that linear regression is invariant at least under invertible morphisms of the predictor variables.

Do we also have naturality under non-invertible linear maps of the predictor space $X$? To answer this, we have to consider the case that the $x_i$ do not together span $X$. In this case, the function $f$ minimizing the error (\ref{eq:linregerr}) is incompletely constrained: there are infinitely many maps $f$ that produce the same minimum value of $E$. The formula (\ref{eq:linregsol}) is also undefined as $\vx^T \vx$ is not invertible. In principle, a learning algorithm could pick one of the many equivalent solutions arbitrarily, but this would not be natural under invertible maps $X\to X$. To show this, it suffices to find an example dataset for which no natural learning algorithm exists. Consider the case that $X=\mathbb{R}^2$, $Y=\mathbb{R}^1$, and there is one training example with predictor $x_1 = \begin{bmatrix}1 \\ 0 \end{bmatrix}$ and target $y_1 = \begin{bmatrix}1 \end{bmatrix}$. Any linear map $f$ encoded by a matrix $\begin{bmatrix}1 & a \end{bmatrix}$ will give the optimum error of $E=0$. Now consider the map $\xi$ encoded by the matrix $\begin{bmatrix} 1 & k \\ 0 & 1 \\ \end{bmatrix}$, for some $k\in \bR$. Then $\xi(x_1)=x_1$, but $f \circ \xi^{-1}$ is encoded by $\begin{bmatrix}1, a-k \end{bmatrix}$. This function is optimal, with an error $E=0$, but it is different to the original (arbitrary) solution $\begin{bmatrix}1,a \end{bmatrix}$. This proves that there is no way of arbitrarily picking amongst the many equally-optimal solutions in an invariant manner. Returning to our original question, we conclude that linear regression is only natural under invertible transformations, as non-invertible transformations will reduce the rank of $\vx$ below $p$.

Finally, we consider naturality in the index set $I$. It is clear from equation (\ref{eq:linregerr}) that the error function does not change on permuting the index set, so we have invariance under permutations. However, we actually have more invariance than this. Let $\va$ be an $M \times N$ matrix with $\va^T \va = 1_N$,  the $N \times N$ identity matrix. (This condition requires $M \geq N$.) If we replace $\vx \mapsto \va \vx$ and $\vx \mapsto \va \vy$, then equation (\ref{eq:linregsol}) shows that $\vf$ is unchanged. Thus, linear regression is invariant not just to permutations of its input examples, but linear recombination of them by an orthogonal projection into a space of possibly higher dimension. In the language of category theory, such maps are known as \emph{monomorphisms} of Euclidean spaces.

We can formalize invariance of linear regression under these transformations within the categorical framework, by changing $\cI$ from a category of sets, to a category of vector spaces. The functor $D$ now defines a dataset $D(X,Y,I)$ as a pair of linear maps: the predictor examples are summarized by a linear map $x:I \to X$, and the targets examples are summarized by a linear map $y:I \to Y$. As before, these maps can be represented by matrices $\vx$ and $\vy$ of size $N\times p$ and $N \times q$, respectively. Now, however, there are a much larger set of morphisms of $I$ that the learning algorithm must respect. The functor $D$ sends a linear map $a:I'\to I$ to a transformation of datasets $D(1_X,1_Y,a):D(X,Y,I)\to D(X,Y,I')$ that sends $x$ to $x \circ a$ and $y$ to $y \circ a$. The argument of the previous paragraph implies that linear regression is a natural transformation between this functor $D$ and the functor $P$. Thus, linear regression is natural in an index category of vector spaces and linear maps satisfying $\va^T \va = 1$, which we will refer to as $\cEuc_{mono}$.

In summary, we have thus shown that linear regression is natural in the categories $\cX=\cFin_{iso}$, $\cY=\cFin$, and $\cI=\cEuc_{mono}$.


\subsection{Ridge regression}

Our second example is ridge regression. This is a variant of linear regression, used when one has high-dimensional inputs and not very many training samples. It avoids overfitting by adding a penalty term to the error function:

\begin{equation}
    \label{eq:ridregerr}
    E = \sum_{i \in I} ||{y_i - f(x_i)}||^2 + \lambda ||f||_{Fro}^2,
\end{equation}

where $||f||_{Fro}^2$ is the Frobenius norm, and $\lambda$ is a parameter of the algorithm. Again, there is an exact solution to this problem. Defining the matrices $\vx$, $\vy$, and $\vf$ as before:

\begin{equation}
\label{eq:ridregsol}
\vf= (\vx^T \vx + \lambda 1_p)^{-1} \vx^T \vy. 
\end{equation}
where $1_p$ denotes the $p \times p$ identity matrix. We no longer require $rank(\vx)=p$ for this solution to hold: there is always a unique function $f$ minimizing the error function.

Ridge regression is natural under the same transformations of $Y$ and $I$ as linear regression, which can be shown by exactly the same arguments. However ridge regression is not natural under arbitrary invertible transformations of $X$. Again, to show this it suffices to find one dataset that demonstrates lack of naturality. Consider a dataset with a single training example, with 1-dimensional predictor and target variables $x_1=b$, $y_1=1$. The ridge regression solution is $f = b/(b^2 + \lambda)$. Naturality under an arbitrary linear transforms of $X$ would require that sending $b \mapsto b c$ would send $f \mapsto f/c$, but the actual value is $bc/(b^2c^2 + \lambda)$, which only equals $f/c$ if $\lambda=0$, i.e. if we are performing ordinary linear regression. 

Instead, ridge regression is natural for $\cX = \cEuc_{mono}$. To show this, consider a transformation $\vx \mapsto \vx \va$, where $a$ is a $p \times r$ matrix $\va$ with $\va \va^T = 1_p$. Algebraic manipulation of equation (\ref{eq:ridregsol}) the transformed predictor $\vf'$ satisfies $\va \vf' = \vf$. The category $\cX$ for which ridge regression is natural is thus neither larger than that the category for linear regression, as it does not include non-orthogonal invertible maps $X\to X$, nor smaller, as it does allow orthogonal maps into a higher-dimensional space.

\section{Summary}

We have described a way to characterize the invariances of a learning algorithm using category theory, and used it to show that linear regression is invariant for predictor variables in the category $\cX=\cFin_{iso}$, while ridge regression is natural for predictor variables in the category $\cX=\cEuc_{mono}$. Both algorithms are invariant for target variables in the category $\cY=\cFin$, and for index variables in the category $\cI=\cEuc_{mono}$. 

This result formalizes and extends the intuitive notion that with linear regression one can rescale the predictor variables to arbitrary units, but with ridge regression the units of measurement cannot be scaled without changing the results. We can draw several other conclusions from the analysis: ridge regression, but not linear regression is invariant under orthogonal transformations into higher-dimensional spaces; and both of them are invariant under arbitrary linear transformations of the targets, and orthogonal rotations of the example space.

The same types of arguments can be applied to any learning algorithms, and extensions to the case of unsupervised learning are also possible. We suggest this framework may be a useful way to characterize invariance of learning algorithms more generally.





\bibliographystyle{unsrt}  
\bibliography{main}  



\section*{Acknowledgements}
This work was supported by the Simons Foundation (325512), Wellcome Trust (205093), ERC (694401), and Chan-Zuckerberg Foundation. 
\end{document}